\documentclass{article} % For LaTeX2e
\usepackage{iclr2024_conference,times}

\usepackage{hyperref}
\usepackage{url}
\usepackage{array} % required for text wrapping in tables

\usepackage{microtype}
\usepackage{graphicx}
\usepackage{subfigure}
\usepackage{booktabs} % for professional tables
\usepackage{bm}
\usepackage{hyperref}
\usepackage[skip=4pt]{caption}

\usepackage{amsmath}
\usepackage{amssymb}
\usepackage{mathtools}
\usepackage{amsthm}
\usepackage{dsfont}
\usepackage[capitalize,noabbrev]{cleveref}
\usepackage[compact]{titlesec}
\titlespacing{\section}{1pt}{*1}{*0.5}
\titlespacing{\subsection}{1pt}{*0.5}{*0}
\theoremstyle{plain}
\newtheorem{theorem}{Theorem}[section]

\theoremstyle{definition}
\newtheorem{definition}[theorem]{Definition}

\theoremstyle{remark}

\usepackage[textsize=tiny]{todonotes}

\newif\ifcomments
\commentstrue 
\ifcomments
    \newcommand{\evef}[1]{{\protect\color{orange}{[EF: #1]}}}
    \newcommand{\jess}[1]{{\protect\color{blue}{[JD: #1]}}}
\else
    \newcommand{\evef}[1]{}
    \newcommand{\jess}[1]{}
\fi

\title{Mapping Social Choice Theory to RLHF}

\author{Jessica Dai* and Eve Fleisig* \\
Department of Computer Science\\
University of California, Berkeley\\
\texttt{\{jessicadai,efleisig\}@berkeley.edu}
}

\iclrfinalcopy % Uncomment for camera-ready version, but NOT for submission.
\begin{document}

\maketitle

\begin{abstract}
Recent work on the limitations of using reinforcement learning from human feedback (RLHF) to incorporate human preferences into  model behavior often raises \textit{social choice theory} as a reference point. Social choice theory's analysis of settings such as voting mechanisms provides technical infrastructure that can inform how to aggregate human preferences amid disagreement. We analyze the problem settings of social choice and RLHF, 
identify key differences between them, and discuss how these differences may affect the RLHF interpretation of well-known technical results in social choice.
% and identify differences between them that prevent well-known technical results in social choice from immediately applying to RLHF. 
We then 
translate % redefine 
canonical desiderata from social choice theory for the RLHF context
% . 
% In addition, we reformulate standard social choice axioms for RLHF 
and discuss how they may serve as analytical tools for open problems in RLHF. 
% Finally, we contextualize the role of social choice in the broader political theory literature on democracy and collective decision making.
\end{abstract}

\section{Introduction and Related Work}
Reinforcement learning from human feedback (RLHF) has recently emerged as a key technique for incorporating human values into AI models.\footnote{ RLHF is the dominant paradigm for incorporating human preferences into language models, but this paper discusses learning human preferences from pairwise or $k$-wise comparisons of outputs more generally. Our analyses are not restricted only to using RL as the method of incorporating these preferences into model behavior.} The central problem setting of RLHF, in which people provide preferences over options that are then used to determine global behavior, shares key similarities with scenarios studied under social choice theory (SCT).\footnote{See \cite{brandt2016handbook} for a textbook treatment of SCT for a computational audience.} Recent work has discussed some of the ways in which SCT can serve as a reference for analysis of RLHF, including direct application of social choice axioms to RLHF \citep{mishra2023ai} and indicating that RLHF implicitly optimizes for the Borda count \citep{Siththaranjan2023DistributionalPL}. Open problems in RLHF identified in surveys by \cite{casper2023open} and \cite{lambert2023history}, among others, include the difficulty of selecting evaluators, accounting for disagreement among evaluators and cognitive biases, non-representative sampled prompts, challenges of developing a single reward function for diverse users, measuring downstream impacts, and assumptions regarding the ease of quantifying or aggregating complex individual preferences. Both \citet{casper2023open} and \citet{ lambert2023history} raise \textit{social choice} as one potential way to analyze some of the problems they identify (see also Appendix \ref{sec:rel-work}). However, differences in the RLHF setting mean that established technical results in SCT require adjustment to be applicable to RLHF. 

In the remainder of this paper, we outline core differences and similarities between the RLHF and SCT problem settings (\S\ref{sec:learning-problems}). We then propose SCT-style axioms for RLHF given the differences in problem settings and discuss open problems raised by these axioms (\S\ref{sec:evaluating}). Finally, we discuss implications of our analysis for conceptualization of 
human preference models in the context of RLHF (\S\ref{sec:discussion}). 
In concurrent work, \cite{conitzer2024social} also advocate for social choice approaches to handling disagreement in RLHF. We propose different formalizations of the problem setting; moreover, their work proposes an axiomatic approach to evaluation, with an emphasis on axioms related to voter behavior. By contrast, we view the axiomatic approach as only one of several relevant approaches, notably with distortion (\S\ref{sec:distortion}) as a key perspective from the social choice literature.

\section{Learning Problems: Preference Modeling and Social Choice}
\label{sec:learning-problems}

We first outline each problem setting and their core differences (see Table \ref{tab:compare}). We notate \textit{alternatives}, the options that voters or evaluators can choose, as $a \in \mathcal A$, with $\mathcal A \subseteq \mathbb R^d$ as the space of all possible alternatives. Each alternative $a$ should be thought of as a \textit{prompt-response pair}, not just a response.\footnote{We choose to consider alternatives at the level of prompt-response pairs, rather than individual responses per prompt. This is because even though RLHF annotations are pairwise comparisons for responses to the same prompt, the high-level goal is to learn a global reward function that can score any prompt-response pair.}
Humans---often called evaluators (preference modeling) or voters (social choice)---are indexed with $i$, with $n$ total evaluators. The set of actual preferences (votes) is denoted $\{\pi\}_{i \in [n]}$, with $\pi_i$, a set of pairwise comparisons, as the preferences from voter $i$. For $a, a' \in \mathcal A$, $a \succ a'$ denotes ``$a$ is preferred to $a'$.''
\begin{table}
    \centering
    \def\arraystretch{1.1}
    \fontsize{9}{10}\selectfont
    \begin{tabular}{l|p{0.315\linewidth}p{0.42\linewidth}}
         &\textbf{Social Choice} & \textbf{RLHF/ Preference Modeling}\\\hline\hline
         \textbf{Space of alternatives} 
         % \\(what are we choosing between?) 
         & Fixed and finite (e.g., candidates in an election) & Evaluators see only samples from a structured but infinite space (e.g. all possible prompts and completions) \\\hline
         \textbf{Inputs to process}
         % \\(what information are we using?)
         &  Fixed set of voters; often assume full information over alternatives (e.g. each voter submits a ballot ranking all candidates) & Set of evaluators (not always representative of general population) give pairwise comparisons over subset of alternatives; no guarantee over which evaluators give feedback on which alternatives \\\hline
         \textbf{Goal of process} 
         % \\(what do we want out of this process?)
         &  A \textit{single winner} (e.g. an elected official) or a set of top-$k$ winners
         & A \textit{reward function} that measures the quality of any alternative, even if unseen (in the election example, a way to quantify not only the goodness of all the current candidates, but also of any new candidate) 
         \\\hline
         \textbf{Evaluation}
         &Preferences of fixed set of voters over fixed alternatives evaluated based on \textit{axioms} (analyze properties of voting rule) or \textit{distortion} (measure utility of outcome)
         & \textit{Generalization}, as measured by accuracy of the reward model or utility/regret of a policy using the reward model, with respect to all possible prompts/completions 
    \end{tabular}
    \caption{\small{Summary of major differences in problem settings of social choice and preference modeling for LLMs.}}
    \label{tab:compare}
\end{table}

\textbf{Human Preference Modeling\quad} 
The goal of human preference modeling\footnote{See \citet{lambert2023history,casper2023open} for detailed surveys, and citations therein for mathematical derivations of the RLHF objective.} is to learn a reward model $r: \mathcal A \to \mathbb R$ that quantifies the desirability of a particular alternative $a$.
The standard data collection method is via pairwise comparisons of text completions for a given prompt, and the standard noise model assumed is a Bradley-Terry model, where the likelihood of \textit{observing} an instance of $a$ being preferred to $a'$ is proportional to how much ``better'' $a$ is than $a'$, i.e.
$
p(a \succ a') = \frac{\exp(r(a))}{\exp(r(a)) + \exp(r(a'))}
$.\looseness=-1 

\textbf{Social Choice\quad} In the standard social choice setting, given a set of alternatives and a set of voters, we receive a ballot from each voter $i$ quantifying the voter's preferences over all $\mathcal{A}$ alternatives. Commonly, this takes the form of \textit{rankings}, where $\pi_i$ gives voter $i$'s rankings of all alternatives. A \textit{voting rule} aggregates all $n$ ballots to produce a single election winner. Voting rules can be evaluated \textit{axiomatically}, i.e. with respect to whether they satisfy particular axioms (properties of the aggregation process), or in terms of \textit{distortion}, i.e. with respect to the (aggregate) benefit derived from selecting the winner over other alternatives.\looseness=-1

\textbf{Properties and goals of the learning problem\quad}
Common assumptions of the social choice problem setting include: that alternatives are \textit{fixed} and finite, i.e. that $\mathcal A$ contains the only options that could possibly be considered;
that we have full information from each voter about each alternative; and that voters and their preferences are fixed (see also Appendix \ref{sec:rel-work}). By contrast, in RLHF, the space of alternatives is structured but infinite. The evaluators, who may not be representative of the full population, give pairwise or $k$-wise preferences over subsets of alternatives, often from a handpicked set of prompts. Each evaluator only sees a small subset of alternatives, and each alternative is only seen by a small subset of evaluators.

In SCT, common goals of the voting process include selecting a single winner (e.g. an elected official) or a set of top-$k$ winners.  By contrast, the goal of RLHF is to produce a reward model that can score the quality of new alternatives. That is, alternatives are not only ranked but assigned real-valued rewards, and the assignment of these rewards must generalize to unseen alternatives.

\subsection{Defining the (Preference Modeling) Voting Rule}

We propose a reformulation of the social choice problem that lets us interpolate between both settings.\footnote{\cite{Siththaranjan2023DistributionalPL} give an interpretation of the maximum-likelihood estimator (MLE) under the Bradley-Terry model as an implementation of Borda count, a standard voting rule from the social choice literature, though without explicitly making the modeling decisions we propose here. 
In future work, we hope to give a deeper analysis of this proposition based on the discussion in Sec. \ref{sec:evaluating}.}

\textbf{Consideration 1: Parameterization of alternatives and preferences.\quad}
First, we assume that alternatives are parameterized in a $d$-dimensional space as $a \in \mathbb R^d$, with $\mathcal A \subseteq \mathbb R^d$. 
Recall that we let $a$ represent a prompt and its completion; accordingly, we continue to assume that all alternatives are commensurable, i.e. that there is a reasonable and well-defined comparison across alternatives.
We also parameterize the preferences of each voter $i$ over text features as $\theta_i \in \mathbb R^d$ and model voter $i$'s reward ascribed to a particular piece of text as $r_{\theta_i}(x) = \langle \theta_i, x\rangle$. In our model, voters have fixed preferences over features, and rewards for each piece of text are \textit{scaled} by those preferences; see also Appendix \ref{sec:rel-work}.

\textbf{Consideration 2: Voters from a population.\quad}
Instead of considering $n$ fixed voters (evaluators), as in social choice, we assume all voters' preferences are drawn from some underlying population $\theta_i \sim \mathcal{V}$.\footnote{Some work in SCT, e.g. the line of work in \citet{dey2015sample}, also considers a population of voters; most, however, assumes fixed voters.} This allows us to model both the scenario in which there exists some shared societal norm from which individual preferences are drawn, and more complicated distributions (e.g. \textit{mixtures} of preference models; \citet{zhao2016learning, zhao2018learning, liu2018efficiently} discuss mixtures of Bradley-Terry and random utility models).
\looseness=-1

With these considerations in mind, we present the preference modeling voting rule. Though the definition can simply be interpreted as a reward model, our statement is intended to emphasize the interplay between how the voting rule aggregates \textit{seen} preferences and how it evaluates \textit{unseen} alternatives.\looseness=-1

\begin{definition}[Preference Modeling Voting Rule]
    A \emph{preference modeling voting rule} $f: \mathbb R^d \to \mathbb R$ is a function that maps some (parameterized) alternative $a \in \mathbb R^d$ to some real-valued score, which represents the population's assessment of the quality of $a$. 
\end{definition}

\section{Evaluating the (Preference Modeling) Voting Rule}
\label{sec:evaluating}

How can social choice inform preference model evaluation? We argue that preference modeling can be divided into two subproblems: that of generalizing a particular set of preferences to new outputs; and that of deciding how to aggregate preferences over outputs. RLHF research has focused almost exclusively on the generalization problem. This perspective, though reasonable under the assumption that evaluators do not meaningfully disagree on their preferences, does not account for meaningful and widespread disagreement between evaluators, as evidenced by work such as \cite{aroyo2023dices}. When evaluators disagree, two perspectives from SCT can help to analyze problems that arise in RLHF: axiomatic approaches and distortion. We discuss the tenets and potential contributions of these three perspectives--generalization, axiomatic approaches, and distortion--in this section.\looseness=-1

\subsection{Perspective 1: Generalization}
Following standard approaches in statistics and machine learning theory, recent work \citep{zhu2023principled, wang2020stretching} studies the problem of estimating $\hat\theta$ under Bradley-Terry models where $r_\theta(x) \propto\langle\theta,x\rangle$.
These approaches take the alternatives to be fixed and predetermined, and all randomness is due to Bradley-Terry. Even under the goal of estimating $\hat\theta$ well (or minimizing the regret of a policy that would use an estimated $\hat\theta$, as analyzed in \citealp{zhu2023principled}), natural extensions could consider the complication of psychological factors that bias observed preferences, such as preferences for ``sycophantic'' text or long outputs \citep{perez-etal-2023-discovering, singhal2023long}; modeling more diverse evaluators; or analyzing the set of prompts to be annotated.\looseness=-1 

\subsection{Perspective 2: Axiomatic Characterizations}

A core tenet of social choice is that voting rules can be \textit{axiomatically} analyzed; i.e., absent the notion of some ``ground truth,'' there are particular principles that the final output should follow. This permits a finer-grained understanding of how aggregations handle individual pieces of input. However, to apply axiomatic analysis to the preference modeling setting, we must still consider ``generalization'' in a sense not considered by standard SCT approaches. 

On a technical level, axioms for the preference modeling setting must satisfy two criteria: (1) they must apply to scores, rather than single winners; and (2) they must distinguish between relationships that apply to the full space of alternatives $\mathcal A$ and all voters, and those that apply only to alternatives and ballots seen at train time. These properties mean that some canonical SCT axioms may be wholly inapplicable, while others may require careful reformulation to apply in the RLHF setting. We highlight examples of each of these cases below.\looseness=-1

\textbf{Example: unanimity, consistency, and Condorcet consistency.\quad}
In the single-winner setting, an alternative $a \in \mathcal A$ is a \textit{Condorcet winner} on a particular set of ballots if, for the majority of voters, $a \succ a'$ for all $a' \neq a$ and $a' \in \mathcal A$. 
A \textit{Condorcet-consistent} voting rule for the single-winner setting always returns the Condorcet winner, if it exists.  \textit{Consistency} is satisfied if when, for every partition of voters, the voting rule selects the same winner, it holds that the voting rule over all the voters also selects that winner. \textit{Unanimity} is satisfied when, if every individual voter expresses the preference $a \succ b$, then the voting rule also selects $a$ over $b$. 

Recall that, for RLHF, 
we are interested not in the final winner but in the scores that $f$ assigns to arbitrary (unseen) alternatives, and that in this setting, arbitrarily small preferences may matter less than differences in reward greater than some margin $\varepsilon$. Also recall that voters are sampled from an underlying population.
To account for these differences, we propose definitions for unanimity, consistency, and Condorcet consistency for the RLHF setting as follows:
% \begin{definition}[(Condorcet?) consistency for preference modeling]
% \label{def:condorcet}
%     First, define $(a, \varepsilon)$-Condorcet consistency as follows. For a fixed $a$ (a potentially unseen alternative), let $\mathcal A'_a \subseteq \mathcal A$ be the set of alternatives such that for all voters, $ \langle\theta_i, (a - a')\rangle > \varepsilon$ for all $a' \neq a$ and $a' \in \mathcal{A}'_a$. Then a (preference modeling) voting rule $f$ is $(a, \varepsilon)$-\text{Condorcet-consistent} when $f(a) - f(a') > \varepsilon$ for all $a' \in \mathcal A'_a$, and $\varepsilon$-\textit{Condorcet-consistent} when it is $(a, \varepsilon)$-\text{Condorcet-consistent} for all $a \in \mathcal A$. If $f$ is $\varepsilon$-Condorcet-consistent for $\varepsilon=0$, then we say $f$ is Condorcet-consistent.
% \end{definition}

\begin{definition}[Unanimity for preference modeling]
\label{def:unanimous}
    First, define $(a, \varepsilon)$-unanimity as follows. For a fixed $a$ (a potentially unseen alternative) and a population of (potentially unseen) voters $\mathcal{V}$, let $\mathcal A'_a \subseteq \mathcal A$ be the set of alternatives such that for all voters, $ \langle\theta_i, (a - a')\rangle > \varepsilon$ for all $a' \neq a$ and $a' \in \mathcal{A}'_a$. Then a (preference modeling) voting rule $f$ is $(a, \varepsilon)$-\text{unanimous} when $f(a) - f(a') > \varepsilon$ for all $a' \in \mathcal A'_a$, and $\varepsilon$-\textit{unanimous} when it is $(a, \varepsilon)$-\text{unanimous} for all $a \in \mathcal A$. If $f$ is $\varepsilon$-unanimous for $\varepsilon=0$, then we say $f$ is unanimous.
\end{definition}

\begin{definition}[Condorcet consistency for preference modeling]
\label{def:condorcet-v2}
    We define $(a, \varepsilon)$-Condorcet consistency as follows. For a fixed $a$ and  voter population $\mathcal{V}$, let $\mathcal A'_a \subseteq \mathcal A$ be the set of alternatives such that $ \mathbb{E}_{\theta_i \sim \mathcal{V}}[\langle\theta_i, (a - a')\rangle] > \varepsilon$ for all $a' \neq a$ and $a' \in \mathcal{A}'_a$. Then a (preference modeling) voting rule $f$ is $(a, \varepsilon)$-\text{Condorcet consistent} when $f(a) - f(a') > \varepsilon$ for all $a' \in \mathcal A'_a$, and $\varepsilon$-\textit{Condorcet consistent} when it is $(a, \varepsilon)$-\text{Condorcet consistent} for all $a \in \mathcal A$. If $f$ is $\varepsilon$-Condorcet consistent for $\varepsilon=0$, then we say $f$ is Condorcet consistent.
\end{definition}

\begin{definition}[Consistency for preference modeling]
\label{def:consistency}
    We define $(a, \varepsilon)$-consistency as follows. For a fixed $a$, let $\mathcal A'_a \subseteq \mathcal A$ be the set of alternatives such that for any sufficiently-large subset of voters $\mathcal{V'} \subseteq \mathcal{V}$, $f_{\mathcal{V'}}(a) - f_{\mathcal{V'}}(a') > \varepsilon$ for all $a' \in \mathcal A'_a$. Then a (preference modeling) voting rule $f$ is $(a, \varepsilon)$-\text{ consistent} when $f_{\mathcal{V}}(a) - f_{\mathcal{V}}(a') > \varepsilon$ for all $a' \in \mathcal A'_a$, and $\varepsilon$-\textit{consistent} when it is $(a, \varepsilon)$-\text{consistent} for all $a \in \mathcal A$. If $f$ is $\varepsilon$-consistent for $\varepsilon=0$, then we say $f$ is consistent.
\end{definition}

Note the distinction between \ref{def:unanimous}, \ref{def:condorcet-v2}, and \ref{def:consistency} lies primarily in \textit{what slice of voters} is being examined; the expectation in \ref{def:condorcet-v2} captures the idea of majority preference, while \ref{def:consistency} is concerned with agreement across subsets of voters. The fact that these axioms can be expressed in terms of which types of \textit{agreement} are important to respect suggests that there is room for more explicit consideration of which types of \textit{dis}agreement are important in preference modeling.

\textbf{Other axioms of (single-winner) social choice may not apply.\quad} \textit{Resolvability} (that the voting rule produces no ties) or \textit{majority} (that any alternative ranked first by a majority of voters should win) have no clear translation to the preference model setting, since choosing a single winner is no longer the main objective. 
\textit{Strategyproofness}---the robustness of the final outcome to strategic behavior by individual voters---is a less pressing concern when the number of alternatives is very large, and when the final outcome is a scoring rule rather than a single winner.\footnote{Even in the context of binary classification, \citet{hardt2023algorithmic} require \textit{coordinated} action of around 10\% of evaluators to substantially affect the output of the learned model.} Despite the seeming inapplicability of some ``classic'' SCT axioms, we argue that it is still worthwhile to consider what new axioms for preference modeling might look like: axioms were developed to establish more fundamental desiderata for democratic processes; moreover, they are designed to apply in cases that lack an obvious ``ground truth'' optimum.\looseness=-1

\subsection{Perspective 3: Distortion}
\label{sec:distortion}
An alternative approach to evaluating voting rules in social choice is through \textit{distortion} (see \cite{anshelevich2021distortion} for a recent survey).
At a high level, distortion is a quantitative notion of \textit{suboptimality} with respect to some information that the voting rule may not be able to access (e.g. the ``hidden context'' of \cite{Siththaranjan2023DistributionalPL}.
In an extension of the standard setting, voters have utilities over alternatives, and submit ballots---rankings---that are consistent with those true utilities. 
\footnote{These utilities are hidden and perhaps unknown even to the voter, so they cannot be elicited directly.} 
The \textit{distortion} of a voting rule is the worst-case difference in utility between the optimal election winner and the winner chosen by the voting rule, where the worst-case is taken over all possible utility functions that would have still been consistent with the observed ballots.

As with axiomatic characterizations, it is nontrivial to redefine distortion directly for RLHF. In SCT, it is reasonable to analyze ``worst-case'' (hidden) utilities due to the finiteness of both the alternatives and the voter, the assumption of no noise in reporting utilities, and because ballots tend to contain information about voters' preferences over the entire space of alternatives. 

For RLHF, therefore, we might consider an approach that captures the conceptual insight of the distortion metric rather than attempting to transliterate it directly. 
For example, we might consider some per-feature weight $w \in \mathbb R^d$ where $w_j$ scales the importance of feature $j$. This could represent phenomena like cognitive bias in labeling, such that annotations are received according to a proxy reward function $r_{\theta, w}(a) = \sum_{j \in [d]} \theta_j w_j a_j$, but where true utilities depend only on $\theta$ and not $w$. On the other hand, this could also reflect the scenario where annotations are given with respect to $r_{\theta} = \langle \theta, a\rangle$ but utility for how accurately a particular feature $\theta_i$ is learned varies by feature (scaled by $w_i$)---for example, someone may prefer both longer text completions and text that avoids gendered stereotypes, but would care more about the final $\theta$ capturing the latter. To operationalize the ``worst-case'' in conjunction with the randomness in voting assumed by Bradley-Terry-Luce, we could work with expected welfare (where the expectation includes randomness in voting); assume some relationship between $\theta$ and $w$; and take the worst-case over possible $\theta$ and $w$ that would be consistent with the observed votes. 
Interestingly, for the statistical inference problem of estimating $\theta$, this is a fundamentally distinct perspective from the standard RLHF approach of maximum likelihood; instead, it is more analogous to estimating a $\theta$ that is robust to ``worst case'' realizations of hidden context.
\looseness=-1

\section{Discussion}
\label{sec:discussion}
SCT provides a rich infrastructure for finer-grained discussion of many difficulties in using RLHF to represent human preferences
by providing theoretical underpinnings for sources of these problems.
These include conceptions of the prompt space, including how prompts for RLHF are chosen and what guarantees on representation may be lacking if the sample is handpicked in an unusual way; and of the evaluator space, including consequences of using an unrepresentative or small set of evaluators. In addition, they may help to conceptualize evaluator behavior: though potentially ``misaligned'' or ``strategic'' evaluators are often discussed \citep{casper2023open}, SCT provides a framework for analyzing the perhaps more common issues of evaluators dealing with poor incentives, incomplete directions, and cognitive biases \citep{singhal2023long, huang2023incorporating}. 
Careful engagement is necessary to combine insights from these communities in a way that is both rigorous and fruitful. One naive interpretation of impossibility results such as Arrow's Theorem is that a version of democracy that relies on direct input from the public is simply untenable.
However, a key aspect of SCT is that the axioms or properties of a voting mechanism that are actually desirable depend greatly on the context of the decision. By focusing on the properties that matter in the RLHF setting, and adapting SCT formulations to differences in the RLHF problem, we reach a space of problems that are both critical and tractable.\looseness=-1 

\subsection*{Acknowledgments}
Many thanks to Bailey Flanigan, Paul Gölz, Nika Haghtalab, Cassidy Laidlaw, and Anand Siththaranjan for their helpful discussions and feedback.

\bibliography{refs}

\begin{thebibliography}{30}
\providecommand{\natexlab}[1]{#1}
\providecommand{\url}[1]{\texttt{#1}}
\expandafter\ifx\csname urlstyle\endcsname\relax
  \providecommand{\doi}[1]{doi: #1}\else
  \providecommand{\doi}{doi: \begingroup \urlstyle{rm}\Url}\fi

\bibitem[Anshelevich et~al.(2018)Anshelevich, Bhardwaj, Elkind, Postl, and Skowron]{ANSHELEVICH201827}
Elliot Anshelevich, Onkar Bhardwaj, Edith Elkind, John Postl, and Piotr Skowron.
\newblock Approximating optimal social choice under metric preferences.
\newblock \emph{Artificial Intelligence}, 264:\penalty0 27--51, 2018.
\newblock ISSN 0004-3702.
\newblock \doi{https://doi.org/10.1016/j.artint.2018.07.006}.
\newblock URL \url{https://www.sciencedirect.com/science/article/pii/S0004370218304569}.

\bibitem[Anshelevich et~al.(2021)Anshelevich, Filos-Ratsikas, Shah, and Voudouris]{anshelevich2021distortion}
Elliot Anshelevich, Aris Filos-Ratsikas, Nisarg Shah, and Alexandros~A Voudouris.
\newblock Distortion in social choice problems: The first 15 years and beyond.
\newblock \emph{arXiv preprint arXiv:2103.00911}, 2021.

\bibitem[Aroyo et~al.(2023)Aroyo, Taylor, Diaz, Homan, Parrish, Serapio-Garcia, Prabhakaran, and Wang]{aroyo2023dices}
Lora Aroyo, Alex~S. Taylor, Mark Diaz, Christopher~M. Homan, Alicia Parrish, Greg Serapio-Garcia, Vinodkumar Prabhakaran, and Ding Wang.
\newblock Dices dataset: Diversity in conversational ai evaluation for safety, 2023.

\bibitem[Benade et~al.(2019)Benade, Procaccia, and Qiao]{Benade2019LowDistortionSW}
Gerdus Benade, Ariel~D. Procaccia, and Mingda Qiao.
\newblock Low-distortion social welfare functions.
\newblock In \emph{AAAI Conference on Artificial Intelligence}, 2019.
\newblock URL \url{https://api.semanticscholar.org/CorpusID:53078427}.

\bibitem[Brandt et~al.(2016)Brandt, Conitzer, Endriss, Lang, and Procaccia]{brandt2016handbook}
Felix Brandt, Vincent Conitzer, Ulle Endriss, J{\'e}r{\^o}me Lang, and Ariel~D Procaccia.
\newblock \emph{Handbook of computational social choice}.
\newblock Cambridge University Press, 2016.

\bibitem[Casper et~al.(2023)Casper, Davies, Shi, Gilbert, Scheurer, Rando, Freedman, Korbak, Lindner, Freire, et~al.]{casper2023open}
Stephen Casper, Xander Davies, Claudia Shi, Thomas~Krendl Gilbert, J{\'e}r{\'e}my Scheurer, Javier Rando, Rachel Freedman, Tomasz Korbak, David Lindner, Pedro Freire, et~al.
\newblock Open problems and fundamental limitations of reinforcement learning from human feedback.
\newblock \emph{arXiv preprint arXiv:2307.15217}, 2023.

\bibitem[Conitzer et~al.(2024)Conitzer, Freedman, Heitzig, Holliday, Jacobs, Lambert, Mossé, Pacuit, Russell, Schoelkopf, Tewolde, and Zwicker]{conitzer2024social}
Vincent Conitzer, Rachel Freedman, Jobst Heitzig, Wesley~H. Holliday, Bob~M. Jacobs, Nathan Lambert, Milan Mossé, Eric Pacuit, Stuart Russell, Hailey Schoelkopf, Emanuel Tewolde, and William~S. Zwicker.
\newblock Social choice for ai alignment: Dealing with diverse human feedback.
\newblock \emph{arXiv preprint arXiv:2404.10271v1}, 2024.

\bibitem[Dey \& Bhattacharyya(2015)Dey and Bhattacharyya]{dey2015sample}
Palash Dey and Arnab Bhattacharyya.
\newblock Sample complexity for winner prediction in elections.
\newblock In \emph{Proceedings of the 2015 International Conference on Autonomous Agents and Multiagent Systems}, pp.\  1421--1430, 2015.

\bibitem[Fish et~al.(2023)Fish, G{\"o}lz, Parkes, Procaccia, Rusak, Shapira, and W{\"u}thrich]{fish2023generative}
Sara Fish, Paul G{\"o}lz, David~C Parkes, Ariel~D Procaccia, Gili Rusak, Itai Shapira, and Manuel W{\"u}thrich.
\newblock Generative social choice.
\newblock \emph{arXiv preprint arXiv:2309.01291}, 2023.

\bibitem[Flanigan et~al.(2023)Flanigan, Halpern, and Psomas]{flanigan2023smoothed}
Bailey Flanigan, Daniel Halpern, and Alexandros Psomas.
\newblock Smoothed analysis of social choice revisited.
\newblock In \emph{International Conference on Web and Internet Economics}, pp.\  290--309. Springer, 2023.

\bibitem[Halpern et~al.(2023)Halpern, Kehne, Procaccia, Tucker-Foltz, and W{\"u}thrich]{halpern2023representation}
Daniel Halpern, Gregory Kehne, Ariel~D Procaccia, Jamie Tucker-Foltz, and Manuel W{\"u}thrich.
\newblock Representation with incomplete votes.
\newblock In \emph{Proceedings of the AAAI Conference on Artificial Intelligence}, volume~37, pp.\  5657--5664, 2023.

\bibitem[Hardt et~al.(2023)Hardt, Mazumdar, Mendler-D{\"u}nner, and Zrnic]{hardt2023algorithmic}
Moritz Hardt, Eric Mazumdar, Celestine Mendler-D{\"u}nner, and Tijana Zrnic.
\newblock Algorithmic collective action in machine learning.
\newblock \emph{arXiv preprint arXiv:2302.04262}, 2023.

\bibitem[Huang et~al.(2023)Huang, Fleisig, and Klein]{huang2023incorporating}
Olivia Huang, Eve Fleisig, and Dan Klein.
\newblock Incorporating worker perspectives into mturk annotation practices for nlp.
\newblock In \emph{Proceedings of the 2023 Conference on Empirical Methods in Natural Language Processing}, pp.\  1010--1028, 2023.

\bibitem[Lambert et~al.(2023)Lambert, Krendl~Gilbert, and Zick]{lambert2023history}
Nathan Lambert, Thomas Krendl~Gilbert, and Tom Zick.
\newblock The history and risks of reinforcement learning and human feedback.
\newblock \emph{arXiv e-prints}, pp.\  arXiv--2310, 2023.

\bibitem[Liu \& Moitra(2018)Liu and Moitra]{liu2018efficiently}
Allen Liu and Ankur Moitra.
\newblock Efficiently learning mixtures of mallows models.
\newblock In \emph{2018 IEEE 59th Annual Symposium on Foundations of Computer Science (FOCS)}, pp.\  627--638. IEEE, 2018.

\bibitem[Liu et~al.(2023)Liu, Zhao, Joshi, Khalman, Saleh, Liu, and Liu]{liu2023statistical}
Tianqi Liu, Yao Zhao, Rishabh Joshi, Misha Khalman, Mohammad Saleh, Peter~J Liu, and Jialu Liu.
\newblock Statistical rejection sampling improves preference optimization.
\newblock \emph{arXiv preprint arXiv:2309.06657}, 2023.

\bibitem[Miller(1992)]{miller1992deliberative}
David Miller.
\newblock Deliberative democracy and social choice.
\newblock \emph{Political studies}, 40\penalty0 (1\_suppl):\penalty0 54--67, 1992.

\bibitem[Mishra(2023)]{mishra2023ai}
Abhilash Mishra.
\newblock Ai alignment and social choice: Fundamental limitations and policy implications.
\newblock \emph{arXiv preprint arXiv:2310.16048}, 2023.

\bibitem[Perez et~al.(2023)Perez, Ringer, Lukosiute, Nguyen, Chen, Heiner, Pettit, Olsson, Kundu, Kadavath, Jones, Chen, Mann, Israel, Seethor, McKinnon, Olah, Yan, Amodei, Amodei, Drain, Li, Tran-Johnson, Khundadze, Kernion, Landis, Kerr, Mueller, Hyun, Landau, Ndousse, Goldberg, Lovitt, Lucas, Sellitto, Zhang, Kingsland, Elhage, Joseph, Mercado, DasSarma, Rausch, Larson, McCandlish, Johnston, Kravec, El~Showk, Lanham, Telleen-Lawton, Brown, Henighan, Hume, Bai, Hatfield-Dodds, Clark, Bowman, Askell, Grosse, Hernandez, Ganguli, Hubinger, Schiefer, and Kaplan]{perez-etal-2023-discovering}
Ethan Perez, Sam Ringer, Kamile Lukosiute, Karina Nguyen, Edwin Chen, Scott Heiner, Craig Pettit, Catherine Olsson, Sandipan Kundu, Saurav Kadavath, Andy Jones, Anna Chen, Benjamin Mann, Brian Israel, Bryan Seethor, Cameron McKinnon, Christopher Olah, Da~Yan, Daniela Amodei, Dario Amodei, Dawn Drain, Dustin Li, Eli Tran-Johnson, Guro Khundadze, Jackson Kernion, James Landis, Jamie Kerr, Jared Mueller, Jeeyoon Hyun, Joshua Landau, Kamal Ndousse, Landon Goldberg, Liane Lovitt, Martin Lucas, Michael Sellitto, Miranda Zhang, Neerav Kingsland, Nelson Elhage, Nicholas Joseph, Noemi Mercado, Nova DasSarma, Oliver Rausch, Robin Larson, Sam McCandlish, Scott Johnston, Shauna Kravec, Sheer El~Showk, Tamera Lanham, Timothy Telleen-Lawton, Tom Brown, Tom Henighan, Tristan Hume, Yuntao Bai, Zac Hatfield-Dodds, Jack Clark, Samuel~R. Bowman, Amanda Askell, Roger Grosse, Danny Hernandez, Deep Ganguli, Evan Hubinger, Nicholas Schiefer, and Jared Kaplan.
\newblock Discovering language model behaviors with model-written evaluations.
\newblock In Anna Rogers, Jordan Boyd-Graber, and Naoaki Okazaki (eds.), \emph{Findings of the Association for Computational Linguistics: ACL 2023}, pp.\  13387--13434, Toronto, Canada, July 2023. Association for Computational Linguistics.
\newblock \doi{10.18653/v1/2023.findings-acl.847}.
\newblock URL \url{https://aclanthology.org/2023.findings-acl.847}.

\bibitem[Procaccia \& Rosenschein(2006)Procaccia and Rosenschein]{procaccia2006distortion}
Ariel~D Procaccia and Jeffrey~S Rosenschein.
\newblock The distortion of cardinal preferences in voting.
\newblock In \emph{International Workshop on Cooperative Information Agents}, pp.\  317--331. Springer, 2006.

\bibitem[Rafailov et~al.(2023)Rafailov, Sharma, Mitchell, Ermon, Manning, and Finn]{rafailov2023direct}
Rafael Rafailov, Archit Sharma, Eric Mitchell, Stefano Ermon, Christopher~D Manning, and Chelsea Finn.
\newblock Direct preference optimization: Your language model is secretly a reward model.
\newblock \emph{arXiv preprint arXiv:2305.18290}, 2023.

\bibitem[Singhal et~al.(2023)Singhal, Goyal, Xu, and Durrett]{singhal2023long}
Prasann Singhal, Tanya Goyal, Jiacheng Xu, and Greg Durrett.
\newblock A long way to go: Investigating length correlations in rlhf.
\newblock \emph{arXiv preprint arXiv:2310.03716}, 2023.

\bibitem[Siththaranjan et~al.(2023)Siththaranjan, Laidlaw, and Hadfield-Menell]{Siththaranjan2023DistributionalPL}
Anand Siththaranjan, Cassidy Laidlaw, and Dylan Hadfield-Menell.
\newblock Distributional preference learning: Understanding and accounting for hidden context in rlhf.
\newblock \emph{ArXiv}, abs/2312.08358, 2023.
\newblock URL \url{https://api.semanticscholar.org/CorpusID:266191810}.

\bibitem[Skowron \& Elkind(2017)Skowron and Elkind]{skowron2017social}
Piotr Skowron and Edith Elkind.
\newblock Social choice under metric preferences: Scoring rules and stv.
\newblock In \emph{Proceedings of the AAAI Conference on Artificial Intelligence}, volume~31, 2017.

\bibitem[Wang et~al.(2020)Wang, Shah, and Ravi]{wang2020stretching}
Jingyan Wang, Nihar Shah, and R~Ravi.
\newblock Stretching the effectiveness of mle from accuracy to bias for pairwise comparisons.
\newblock In \emph{International Conference on Artificial Intelligence and Statistics}, pp.\  66--76. PMLR, 2020.

\bibitem[Wilde et~al.(2022)Wilde, Biyik, Sadigh, and Smith]{wilde2022learning}
Nils Wilde, Erdem Biyik, Dorsa Sadigh, and Stephen~L Smith.
\newblock Learning reward functions from scale feedback.
\newblock In \emph{Conference on Robot Learning}, pp.\  353--362. PMLR, 2022.

\bibitem[Wu et~al.(2023)Wu, Hu, Shi, Dziri, Suhr, Ammanabrolu, Smith, Ostendorf, and Hajishirzi]{wu2023fine}
Zeqiu Wu, Yushi Hu, Weijia Shi, Nouha Dziri, Alane Suhr, Prithviraj Ammanabrolu, Noah~A Smith, Mari Ostendorf, and Hannaneh Hajishirzi.
\newblock Fine-grained human feedback gives better rewards for language model training.
\newblock \emph{arXiv preprint arXiv:2306.01693}, 2023.

\bibitem[Zhao et~al.(2016)Zhao, Piech, and Xia]{zhao2016learning}
Zhibing Zhao, Peter Piech, and Lirong Xia.
\newblock Learning mixtures of plackett-luce models.
\newblock In \emph{International Conference on Machine Learning}, pp.\  2906--2914. PMLR, 2016.

\bibitem[Zhao et~al.(2018)Zhao, Villamil, and Xia]{zhao2018learning}
Zhibing Zhao, Tristan Villamil, and Lirong Xia.
\newblock Learning mixtures of random utility models.
\newblock In \emph{Proceedings of the AAAI Conference on Artificial Intelligence}, volume~32, 2018.

\bibitem[Zhu et~al.(2023)Zhu, Jiao, and Jordan]{zhu2023principled}
Banghua Zhu, Jiantao Jiao, and Michael~I. Jordan.
\newblock Principled reinforcement learning with human feedback from pairwise or $k$-wise comparisons, 2023.

\end{thebibliography}
\bibliographystyle{iclr2024_conference}

\newpage
\appendix
\section{Additional Related Work}
\label{sec:rel-work}

\paragraph{Social choice}

Instead of applying social choice to AI, \citet{fish2023generative} show that applying AI to augment democratic processes can improve canonical social choice results. In line with our discussion of application-specific reworkings of social choice axioms, \cite{flanigan2023smoothed} uses smoothed analysis for relaxation of worst-case social choice axioms, which helps to distinguish voting rules that rarely satisfy axioms from those that often do.

\cite{halpern2023representation} discuss relaxed assumptions on fixed preferences and full information. \cite{fish2023generative} discusses relaxations of assumptions regarding fixed, finite, and commensurable preferences. The literature on \textit{metric preferences} (see e.g. \citet{skowron2017social, ANSHELEVICH201827}) also parameterizes alternatives in Euclidean space; however, voters' preferences over them are determined by their \textit{distance} from each alternative.

In single-winner elections, the distortion of Borda count is unbounded \citep{procaccia2006distortion}; however, numerical experiments from \citet{Benade2019LowDistortionSW} suggest that Borda count may be near-optimal in a setting where the voting rule outputs a ranking over alternatives rather than a single winner.

\paragraph{RLHF}

 Related empirical RLHF work has raised alternative ways to collect human preferences, such as measuring the degree of preference \citep{wilde2022learning} or soliciting fine-grained preferences that compare alternatives along multiple dimensions \citep{wu2023fine}.
Alternative optimization methods to RLHF have also been proposed (e.g. \citet{rafailov2023direct, liu2023statistical}).

\paragraph{Political theory}

For democratic theorists, social choice results over the last few decades have provided both challenges and clarity to normative discussions of what makes collective decision-making ``legitimate.''
For example, \textit{types} of choices have different normative consequences (e.g., choosing the best way to explain a math problem or the best way to discuss a conspiracy theory), which links to discussion of what decisions are best suited to different voting rules; for instance, \citet{miller1992deliberative} argue that Borda count often works well, but key normative decisions might demand majoritarianism. In the context of LMs, this distinction also helps to distinguish between \textit{preferences} that can be personalized and universal \textit{judgments} that must affect all users.

\end{document}